\documentclass[10pt]{article}

\usepackage[utf8]{inputenc}
\usepackage{amsmath}
\usepackage{amsfonts}
\usepackage{mathrsfs}
\usepackage{amssymb}
\usepackage{amsthm}
\usepackage{geometry}
\usepackage{libertine}
\usepackage{wrapfig}
\usepackage{graphicx}
\usepackage{multirow}
\usepackage{epstopdf}
\usepackage{multicol}
\usepackage{subfigure}
\usepackage{booktabs}

\usepackage{algorithm}
\usepackage{algorithmic}
\usepackage[algo2e,linesnumbered]{algorithm2e}
\usepackage{amssymb}
\usepackage{tabularx}
\usepackage{appendix}
\usepackage{natbib}

\usepackage[table]{xcolor}
\definecolor{mydarkred}{rgb}{0.6,0,0}
\definecolor{mydarkgreen}{rgb}{0,0.6,0}
\usepackage[colorlinks,
linkcolor=mydarkred,
citecolor=mydarkgreen]{hyperref}
\usepackage{url}
\usepackage{bm}
\usepackage{pifont}

\title{Extended T: Learning with Mixed Closed-set and Open-set Noisy Labels}

\author{
  Xiaobo Xia$^{1}$,
  Tongliang Liu$^{1}$,
  Bo Han$^{2}$,\\
  Nannan Wang$^3$,
  Jiankang Deng$^4$,
  Jiatong Li$^5$,
  Yinian Mao$^5$\\
  $^1$University of Sydney;
  $^2$Hong Kong Baptist University;\\
  $^3$Xidian University;
  $^4$Imperial College London;
  $^5$Meituan\\
}
\date{}
\begin{document}
\bibliographystyle{plainnat}

\maketitle

\begin{abstract}
The label noise transition matrix $T$, reflecting the probabilities that true labels flip into noisy ones, is of vital importance to model label noise and design statistically consistent classifiers. The traditional transition matrix is limited to model closed-set label noise, where noisy training data has true class labels within the noisy label set. It is unfitted to employ such a transition matrix to model open-set label noise, where some true class labels are outside the noisy label set. Thus when considering a more realistic situation, i.e., both closed-set and open-set label noise occurs, existing methods will undesirably give biased solutions. Besides, the traditional transition matrix is limited to model instance-independent label noise, which may not perform well in practice. In this paper, we focus on learning under the mixed closed-set and open-set label noise. We address the aforementioned issues by extending the traditional transition matrix to be able to model mixed label noise, and further to the cluster-dependent transition matrix to better approximate the instance-dependent label noise in real-world applications. We term the proposed transition matrix as the cluster-dependent extended transition matrix. An unbiased estimator (i.e., extended $T$-estimator) has been designed to estimate the cluster-dependent extended transition matrix by only exploiting the noisy data. Comprehensive synthetic and real experiments validate that our method can better model the mixed label noise, following its more robust performance than the prior state-of-the-art label-noise learning methods.
\end{abstract}

\vspace{15pt}
\section{Introduction}
The success of deep networks largely relies on large-scale datasets with high-quality label annotations \cite{han2018masking,wang2018iterative,jiang2020beyond}. However, it is quite costly, time-consuming, or even infeasible to collect such data. Instead, in practice, many large-scale datasets are collected by cheap ways, e.g., from search engines or web crawlers. The obtained data contains label noise \cite{yao2020searching,wang2020fair}. The presence of label noise may adversely affect the model prediction and generalization performance \cite{zhang2017understanding}. It is therefore of great importance to train deep networks robustly against label noise.

\begin{figure}[t] 
\centering
\includegraphics[width=0.40\textwidth]{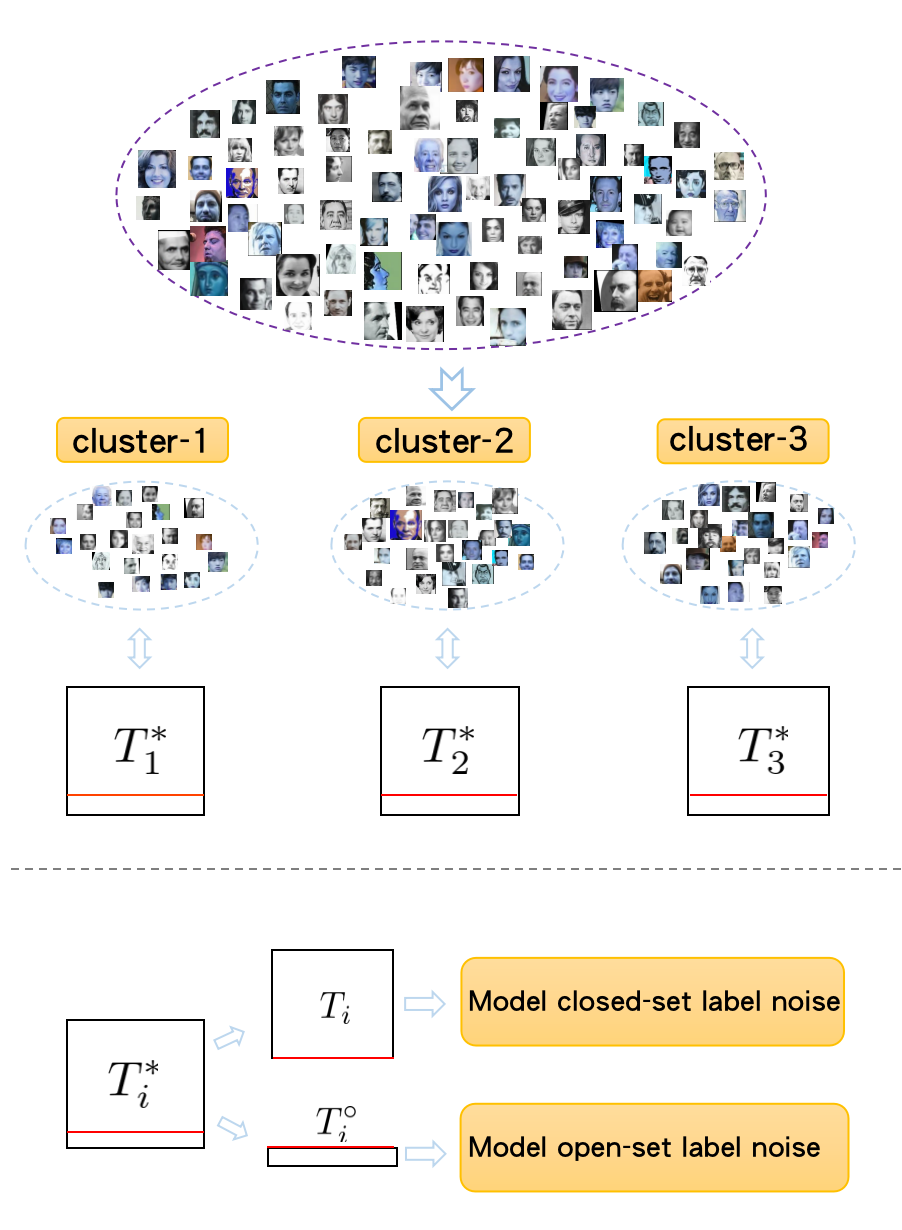}
\caption{Illustration of the cluster-dependent extended transition matrix. The face images are collected from MS1MV0 \cite{guo2016ms}. The proposed method learns the transition matrix for different clusters and extends the traditional transition matrix to be able to model mixed label noise.} 
\label{fig:main} 
\end{figure}

The types of label noise studied so far can be fell into two categories: \textit{closed-set} and \textit{open-set} label noise. The closed-set label noise occurs when instances with noisy labels have true class labels within the noisy label set \cite{wang2018iterative}. Oppositely, the open-set label noise occurs when instances with noisy labels have some true class labels outside the noisy label set \cite{wang2018iterative}. 
The closed-set label noise has been extensively studied, e.g., \cite{liu2016classification,han2018masking,zheng2020error,yao2020dual,menon2020can,li2020dividemix,nguyen2020self,liu2019peer,liu2020early,han2020sigua,wei2020optimizing,cheng2020learning}. There are also some pioneer work studying the open-set label noise, e.g., \cite{wang2018iterative,zhang2018generalized,yu2019does}. All these methods are designed for handling the closed-set and open-set label noise independently and cannot handle the mixed closed-set and open-set label noise well. However, it is more practical that the two types of label noise co-occur in real-world applications \cite{sachdeva2020evidentialmix}. For example, many large-scale face recognition datasets are automatically collected via image search engines. The face images in the datasets contain both two types of label noise \cite{wang2019co}. 

One promising strategy for combating label noise is to model label noise. Compared with model-free methods which empirically work well but does not model the label noise explicitly, the reliability of model-based methods is better guaranteed \cite{patrini2017making,xia2020parts,berthon2020idn}. By utilizing the \textit{transition matrix} which denotes the probabilities that clean labels ﬂip into noisy ones, model-based methods have been verified to be able to deal with closed-set label noise, mainly with the kind of class-dependent (instance-independent) label noise \cite{cheng2017learning,natarajan2013learning,shu2020meta}. However, the traditional model-based methods have the following limitations. Firstly, they cannot model open-set label noise and will provide \textit{biased} solutions when there exists mixed closed-set and open-set label noise. Secondly, the instance-dependent label noise is common in real-world applications as difficult instances are prone to have inaccurate labels. It is \textit{ill-posed} to 
learn the instance-dependent transition matrix by only exploiting the noisy training data as discussed in \cite{xia2020parts}. Class-dependent transition matrix has been exploited to approximate the instance-dependent transition matrix. However, the approximation error is large when the label noise rate is high.

In this paper, we present a novel method for learning under the mixed closed-set and open-set label noise. The proposed method extends the traditional transition matrix to be able to model the mixed label noise and better approximate the instance-dependent label noise. Specifically, we integrate all open-set classes as a \textit{meta class}, which is paratactic with the other true classes in the closed set. We then identify \textit{anchor points} belonging to the meta class of the open-set and the true classes of the closed set. The extended transition matrix involving the meta class can be unbiasedly estimated by exploiting anchor points. To further handle the instance-dependent label noise in reality, we exploit cluster-dependent transition matrices to better approximate the instance-dependent transition matrix. Specifically, we divide all training examples into several clusters (with the constraint that cluster contains anchor points for the meta class of the open set and true classes of the closed set). The cluster-dependent transition matrix can then be unbiasedly estimated for each cluster. The cluster-dependent transition matrices capture the geometric information of instances and thus can better approximate the instance-dependent transition matrix than the class-dependent transition matrix. The illustration of the proposed method is provided in Fig.~{\ref{fig:main}}.

The main contributions of this paper are three-fold:
\begin{itemize}
	\item We focus on learning under the mixed closed-set and open-set label noise, and extend the traditional transition matrix to be able to model the mixed label noise. 
	\item We propose the cluster-dependent extended transition matrices to handle instance-dependent label noise in real-world applications, which produces a more reliable solution.
	\item We conduct comprehensive experiments on synthetic and real-world label-noise datasets to demonstrate that the proposed method achieves superior robustness over the baselines under the mixed label noise. 

\end{itemize}

The rest of the paper is organized as follows. In Section \ref{sec2} , we review related works on learning with label noise. In section \ref{sec3}, we introduce some notations and background knowledge. In Section \ref{sec4}, we introduce the proposed method in details. Experimental results and analyses are provided in Section \ref{sec5}. Finally, we conclude the paper in Section \ref{sec6}.

\section{Related Work}\label{sec2}
\textbf{Learning with the noise transition matrix.} The label noise transition matrix plays an essential role in modeling the label noise, which reﬂects the probabilities that true labels flip into other noisy ones \cite{natarajan2013learning,scott2015rate,hendrycks2018using,wu2020class2simi}. For learning under closed-set label noise, where true classes are known classes in the training data, the noise transition matrix can effectively infer the \textit{clean class posterior probability} with the \textit{noisy class posterior probability} \cite{liu2016classification}. Thus, we can assign clean labels for given instances, even though only noisy training data are available \cite{patrini2017making}. Many advanced methods borrow this idea and learn the noise transition matrix to handle label noise \cite{liu2016classification,shu2020meta}. For example, for estimating the class-dependent (closed-set label noise) transition matrix more accurately, a slack variable can be introduced to revise the initialized transition matrix \cite{xia2019revision}. An intermediate class can be used to avoid directly estimating the noisy class posterior \cite{yao2020dual}. For modeling the instance-dependent closed-set label noise more accurately, the label noise can be assumed to depend only on the parts of instances \cite{xia2020parts}. For learning under open-set label noise, true classes of noisy training data are \textit{outside} the set of known classes. Recall the definition of the (traditional) transition matrix, the flip probabilities indicate the rates of the true classes flipped to the noisy ones. If we use the method of modeling closed-set label noise to model open-set label noise, we will mistakenly treat some unknown incorrect classes as true classes, which leads to poor classification performance. To the best of our knowledge, how to effectively model the mixed closed-set and open-set label noise is a new challenge. \\
\textbf{Other methods of learning under label noise.} We first briefly introduce other algorithms for dealing with closed-set and open-set label noise separately without modeling the noise explicitly. For example, some approaches to handle label noise are to exploit the memorization effects of deep models and extract reliable examples with small loss from the training data \cite{jiang2018mentornet,han2018co,yu2019does}. Some approaches to handle label noise are to reweight examples \cite{ren2018learning,wang2018iterative,shu2019meta}. Some works aim to design robust loss functions to cope with label noise \cite{wang2019symmetric,menon2020can,liu2019peer}. Some other works (implicitly) add regularization \cite{guo2018curriculumnet,veit2017learning,vahdat2017toward,jiang2020beyond,lukasik2020does}. Then, we introduce the pioneer work for dealing with mixed closed-set and open-set label noise without modeling the noise explicitly. EvidentialMix \cite{sachdeva2020evidentialmix} focuses on synthetic mixed label noise and achieves promising performance by combining DivideMix \cite{li2020dividemix} and SL loss \cite{sensoy2018evidential}. Note that our work is fundamentally different from EvidentialMix. The main reasons are as follows. (1) EvidentialMix combines several of advanced approaches, but our work focuses on one, i.e., learning with the noise transition matrix. (2) EvidentialMix works well, but does not model the mixed label noise. Our work models the mixed label noise and improves the reliability of the method. We suggest that the readers can refer \cite{han2020survey} for more methods of learning under label noise. \\
\textbf{Deep clustering.} As a typical unsupervised learning method, clustering has been widely used in various tasks. It aims to keep similar data points in the same cluster while dissimilar ones in different clusters. Clustering is proven to be able to find representative data points among all data points \cite{wang2008clustering,teichgraeber2019clustering}. Benefit from the power of deep learning, lots of approaches boost traditional clustering techniques, e.g., $k$-means \cite{jain2010data} and spectral clustering \cite{ng2002spectral}, by using deep models \cite{li2018discriminatively,yang2016joint}. They cluster on deep representations instead of original features as deep representations have a higher degree of discrimination \cite{yang2019deep}.

\section{Preliminaries}\label{sec3}
In this section, we introduce the notation, problem setting and background knowledge for the label noise transition matrix. 

In this paper, we consider the $c$-class classification problem. Let $\mathcal{X}$ and $\mathcal{Y}$ be the input and output domains, where $\mathcal{Y}=\{1,\ldots,c\}$ is the label set. Let $D$ be the clean joint distribution of a pair of random variables $(X,Y)\in\mathcal{X}\times\mathcal{Y}$, and $\{(x_i,y_i)\}_{i=1}^n$ be a training sample drawn from $D$. 
 Let $\Tilde{D}$ be the noisy joint distribution of the noisy random variables $(X,\Tilde{Y})\in\mathcal{X}\times\mathcal{Y}$.
In learning under the mixed label noise, what we can access is a noisy training sample $\{(x_i,\Tilde{y}_i)\}_{i=1}^n$ drawn from a noisy distribution $\Tilde{D}$, where $\Tilde{y}$ is the possibly corrupted label of the underlying clean label $y$.  Our aim is to learn a robust classifier from the noisy sample that can assign clean labels for test instances.  

We exploit a deep neural network to train our classifier. The network comprises a transformation $h:\mathcal{X}\rightarrow\mathbb{R}^c$, where $h=h^{(d)}\circ h^{(d-1)}\circ\cdots\circ h^{(1)}$ is the composition of a series of transformation layers $h^{(i)}$. Specifically, 
\begin{align*}
    &h^{(i)}(z)=\sigma(W^{(i)}z+b^{(i)}), i=1,\ldots,d-1,\\
    &h^{(d)}(z)=W^{(d)}z,
\end{align*}
where $W^{(i)}$ and $b^{(i)}$ are parameters to be learned, and $\sigma$ is the activation function, e.g., Sigmoid \cite{yin2003flexible} and ReLU \cite{glorot2011deep}. Note that the outputs after the layer $h^{(d-1)}$ can be regarded as the \textit{deep representations} of the instances. For the classification task, the output layer $f$ is always set as a \textit{softmax} layer to imitate classificaition probabilities, i.e., $f_i(x)=\text{exp}(h_i(x))/\sum_{k=1}^c\text{exp}(h_k(x))$. Namely, $f(X)=\hat{P}(Y|X)$. For traditional supervised learning (the classification task), the optimal classifier is defined by that one that minimizes the expected risk: 
\begin{align}\label{eq:expect_risk}
    R_{\ell,D}(f)=\mathbb{E}_{(X,Y)\sim D}[\ell(f(X),Y)],
\end{align}
where $\mathbb{E}_{(X,Y)\sim D}$ denotes the expectation over the distribution $D$, and $\ell:\mathbb{R}^c\times\mathcal{Y}\rightarrow\mathbb{R}_{+}$ is a surrogate loss function for $c$-class classification, e.g., the \textit{cross-entropy loss}. As $D$ is usually unknown, the empirical risk has been exploited as follows to approximate the expected risk:
\begin{align}\label{eq:empirical_risk_clean}
    R_{\ell,n}(f)=\frac{1}{n}\sum_{i=1}^n\ell(f(x_i),y_i).
\end{align}
However, in learning under label noise, the clean training sample is not available.
We only have the access to the noisy training sample. By exploiting the label noise transition matrix, the following modified empirical risk has been proposed to well approximate $R_{\ell,D}(f)$:
\begin{align}\label{eq:empirical_risk_noisy}
    \Tilde{R}_{\tilde{\ell},n}(f)=\frac{1}{n}\sum_{i=1}^n\tilde{\ell}(f(x_i),\Tilde{y}_i),
\end{align}
where $\tilde{\ell}$ represents the modified loss function by exploiting the transition matrix. It has been proven that the empirical risk $\Tilde{R}_{\tilde{\ell},n}(f)$ will converge to the expected risk  $R_{\ell,D}(f)$ if the transition matrix can be unbiasedly estimated \cite{liu2016classification,patrini2017making,yao2020dual}. Then the corresponding learned classifier will be statistically consistent because it will converge to the optimal one defined by the clean data.

In this work, we will study how to use the transition matrix to model the mixed closed-set and open-set label noise, based on which, robust classifiers can be designed. We first formally introduce the traditional transition matrix, i.e., $T\in[0,1]^{c\times c}$, which is only capable of modeling the closed-set label noise. The transition matrix generally depends on the instances and the true labels, i.e., $T_{ij}(x)=P(\Tilde{Y}=j|Y=i,X=x)$ \cite{cheng2017learning}. Unfortunately, the instance-dependent transition matrix is unidentifiable without any assumption \cite{xia2019revision,xia2020parts}. The existing methods approximate the instance-dependent transition matrix by assuming that the noise transition matrix is \textit{class-dependent} and \textit{instance-independent}, i.e., $T(x)=P(\Tilde{Y}=j|Y=i,X=x)=P(\Tilde{Y}=j|Y=i)$. When there is no confusion, we will short-hand $T(x)$ as $T$ for the class-dependent transition matrix. Note that the noisy class posterior $P(\Tilde{Y}|X)$ can be estimated by using the noisy training data. With the transition matrix, we can bridge the noisy class posterior $P(\Tilde{Y}|X)$ and the clean class posterior $P(Y|X)$ as follows: 
\begin{align}\label{eq:transition}
    P(\Tilde{Y}=j|X=x) = \sum_{i=1}^cT_{ij}P(Y=i|X=x).
\end{align}
The matrix form of Eq.~(\ref{eq:transition}) can be written as $P(\Tilde{Y}|X)=T^\top P(Y|X)$. Namely, when learning under label noise, if we have the access to the ground-truth transition matrix, we can recover $P(Y|X)$ with $P(\Tilde{Y}|X)$. The transition matrix can be therefore used to build \textit{classifier-consistent} or \textit{risk-consistent} algorithms \cite{liu2016classification,xia2019revision}. 
However, the traditional transition matrix fails to handle the mixed closed-set and open-set label noise as it encodes no open-set class information. We will discuss how to extend it to better handle the instance-dependent mixed closed-set and open-set label noise in the next section.

\section{Methodology}\label{sec4}
 In this section, we first show how to model the mixed label noise by extending the traditional label noise transition matrix.
 Then we present how to learn the cluster-dependent extended transition matrices for better approximating the instance-dependent mixed label noise. Finally, we show how to exploit the extended transition matrix for training a robust classifier.

\subsection{Class-dependent extended T}\label{sec:4.1}
As discussed in Section \ref{sec3}, the traditional label noise transition matrix is a $c\times c$ matrix linking the noisy class information to the closed-set clean class information without considering the open-set clean class information. It is therefore limited to handle the open-set label noise problems. Taking the traditional class-dependent transition matrix as an example, we  discuss how to extend it to handle the open-set label noise. Note that in the next subsection, we will discuss how to handle the instance-dependent mixed label noise.

To model the open-set label noise, we introduce a \textit{meta class} which is an integration of all the possible open-set classes. As shown in Figure \ref{fig:main}, we extend the traditional transition matrix to $(c+1)\times c$ dimensional, where the additional $1\times c$ vector denoted by $T^{\circ}$ represents how the meta class (or the open-set class) flips into the closed-set classes, i.e., $P(\tilde{Y}=j|Y=m,X=x)$, where $j=1,...,c$ and $m$ represents the meta class label. The extended transition matrix encodes the open-set class information and thus can be exploited to better reduce the side-effect of the open-set label noise.

We then discuss how to estimate the extended transition matrix by exploiting \textit{anchor points}.
Anchor points are defined in the clean data domain \cite{liu2016classification}. Formally, an instance $x$ is an anchor point for the class $i$ if $P(Y=i|X=x)$ is equal to one or approaches one. Given an anchor point $x$, if $P(Y=i|X=x)=1$, we have that for $k\neq i$, $P(Y=k|X=x)=0$. Then, we have, 
\begin{equation}\label{eq:anchor}
	P(\Tilde{Y}=j|X=x)=\sum_{k=1}^cT_{kj}P(Y=k|X=x)=T_{ij}.
\end{equation}
The equation holds because we assume that the transition matrix is class-dependent and instance-independent, e.g., $T_{ij}(x)=T_{ij}$.

Therefore, the transition matrix $T$ can be unbiasedly estimated via estimating the noisy class posteriors for the anchor point of each class (including the meta class). Note that the anchor point assumption is a widely adopted in the literature of learning with noisy labels \cite{liu2016classification,yao2020dual,yu2018learning}. We could follow them and assume the availability of anchor points. 

However, assuming that anchor points are given could be strong for many real-world applications. We could release the assumption by just assuming that the anchor points exist in the training data and then design algorithms to locate them. By the definition of anchor points, they are the representatives of each class. 
The $k$-means cluster technique \cite{jain2010data} therefore could be utilized to detect anchor points. Note that to fulfil the power of deep learning, we could make use of the deep features to find anchor points. Note that even if the training labels are noisy, the deep features obtained by exploiting a deep neural network and the noisy training sample still capture the class-discriminative information and could be exploited for finding anchor points. (Note that the obtained deep features are not sufficient for learning the class posteriors.)

Specifically, for the training example $(x_1,\ldots,x_n)$, the number of clusters is set to $k=c+1$. The aim is to detect the anchor points of the $c$ classes in the closed set and the meta class in the open set. 
For $k$-means, we formulate the loss function as: 
\begin{equation}
    \ell_k=\sum_{i=1}^n\sum_{k=1}^{c+1}M_{ik}\|x_i-\mu_k\|_2^2, 
\end{equation}
where $M$ is the cluster matrix, $M_{ik}=1$ if $x_i$ belongs to the $k$-th cluster, otherwise $M_{ik}=0$. The symbol $\mu_k$ represents the $k$-th cluster centroid \cite{jain2010data}. By assuming that the open-set examples are of minority, the top $c$ largest clusters are selected for the closed-set classes and the remaining cluster is deemed as the meta-class cluster. To assign class labels to the closed-set clusters, we first count the noisy class labels in each cluster. Then we regard the cluster with the most examples of a certain noisy class $i$ as the $i$-class cluster. When detecting the anchor points belonging to the closed-set classes, we follow \cite{patrini2017making,xia2019revision}. Then, we use these anchor points to estimate the transition matrix $T$ for modeling the closed-set label noise with Eq.~(\ref{eq:anchor}). 

For detecting the anchor points belonging to the meta class, we determine that the anchor points are the data points which are close to the centroid of the meta class cluster. This way of detection is reasonable. Recall the definition of anchor points (i.e., data points that belong to a speciﬁc class almost surely), it reveals that anchor points are the most representative data points. Clustering has been verified to be able to detect the the representative data points. The cluster centroid is the representative of the cluster. Thus for the meta class, we intuitively set the data point close to the centroid as the anchor point. Then with Eq.~(\ref{eq:anchor}), we use the noisy class posterior probabilities of the anchor point to estimate the transition matrix for modeling the open-set label noise. We denote the transition matrix for open-set label noise  as $T^{\circ}\in[0,1]^{1\times c}$. When finish the estimation of the transition matrix for both types of label noise, we combine $T$ and $T^{\circ}$ to obtain the extended transition matrix $T^{\star}\in[0,1]^{(c+1)\times c}$ to model the mixed label noise. 

\subsection{Cluster-dependent extended T}\label{sec:4.2}
We have presented how to model the mixed label noise by using class-dependent extended transition matrix and how to estimate the extended transition matrix. However, many real-world label noise is instance-dependent. To handle this problem, we propose to use cluster-dependent extended transition matrices to better model the instance-dependent label noise, which is based on the intuition that the instances which have similar features are more prone to have the similar label flip process \cite{cheng2017learning}. We thus can employ the same extended transition matrix to model the mixed label noise for the instances which have similar features. We term such extended transition matrices as cluster-dependent extended transition matrices.

We show how to learn the cluster-dependent transition matrices as follows. Consider the training examples $(x_1,\ldots,x_n)$, we cluster on their deep representations again to obtain clusters. The total number of the clusters is set to a small number, i.e., ${k}$. Note that when we set the value of ${k}$ for clustering, we have to ensure that there are anchor points for each classes in each cluster. The overall procedure to learn the cluster-dependent  extended transition matrices is summarized in Algorithm \ref{alg:main}.

\begin{algorithm}[h]
{\bfseries Input}: Noisy training sample $\mathcal{D}_t$, noisy validation sample $\mathcal{D}_v$, the number of cluster-dependent transtion matrices $k$;\\
1: Train a deep model by using the noisy data $\mathcal{D}_t$ and $\mathcal{D}_v$;\\
2: Get the deep representations of the examples by employing the trained deep network;\\
3: Detect the anchor points used for estimation with clustering; \\
4: Cluster on the deep representations of the examples to obtain $k$ clusters;\\
\For{$i=1,\dots,k$}{
	
	5: Estimate the transition matrix $T_i$ for the closed-set label nosie;\\
	6: Estimate the transition matrix $T_i^\circ$ for the open-set label noise;\\
	7: Obtain the cluster-dependent transition matrix $T_i^{\star}$;\\ 	
}
8: {\bfseries Output}: $T_1^{\star},\ldots,T_{k}^{\star}$.
\caption{Cluster-dependent Transition Matrices Learning Algorithm}
\label{alg:main}
\end{algorithm}

\subsection{Learning with importance reweighting}\label{sec:4.3}
In previous subsection, we have presented the methods about how to learn the cluster-dependent extended transition matrices to model the mixed instance-dependent label noise. With the learned extended transition matrix $T_i^{\star} (i=1,\ldots,{k})$, we employ the \textit{importance reweighting} technique \cite{gretton2009covariate,liu2016classification,fang2020rethinking} to train a robust classifier against the label noise. Importance reweighting can be exploited to rewrite the expected risk w.r.t. clean data (Eq.~(\ref{eq:expect_risk})) as follows: 
\begin{eqnarray*}
&R_{\ell,D}(f)&=\mathbb{E}_{(X,Y)\sim D}\left[\ell(f(X),Y)\right]\\
&&=\mathbb{E}_{(X,\Tilde{Y})\sim \Tilde{D}}\left[\frac{P_{D}(X,Y)}{P_{\Tilde{D}}(X,\Tilde{Y})}\ell(f(X),\Tilde{Y})\right]\\
&&=\mathbb{E}_{(X,\Tilde{Y})\sim \Tilde{D}}\left[\beta(X,\Tilde{Y})\ell(f(X),\Tilde{Y})\right]\\
&&=\mathbb{E}_{(X,\Tilde{Y})\sim \Tilde{D}}\left[\bar{\ell}(f(X),\Tilde{Y})\right].
\end{eqnarray*}
If $\Tilde{Y}=i$, $\bar{\ell}(f(X),\Tilde{Y})=\bar{\ell}(f(X),i)=\frac{P_D(Y=i|X=x)}{P_{\Tilde{D}}(\Tilde{Y}=i|X=x)}\ell(f(X),i)$. As $P_{\Tilde{D}}(\Tilde{Y}=i|X=x)\neq 0$ during training, the importance reweighting method is stable without truncating the importance ratios \cite{liu2016classification}. 
For the $c$-class classification problem under the mixed label noise, by exploiting the cluster-dependent extended transition matrices $T^{\star}$ (we hide the index for simplify), the empirical risk can be formulated as: 
\begin{align}
\label{eq:em_importance}%
{
\tilde{R}_{\tilde{\ell},n}=\frac{1}{n}\sum_{i=1}^{n}\frac{{g}_{\Tilde{y}_i}(x_i)}{(T^{\star\top}{g})_{\Tilde{y}_i}(x_i)}\ell(f(x_i),\Tilde{y}_i),}
\end{align}
where $f(x)=(T^{\star\top}{g})(x)$ and $g(x)$ is the output of the softmax layer. We use $\arg\max_{j\in\{1,\ldots,c\}}{g}_j(x)$ to assign labels for the test data. Note that during training, the $T^{\star}$ is determined according to the cluster to which the example $x_i$ belongs. As we detect the anchor points from the noisy training data, as did in \cite{xia2019revision,xia2020parts,patrini2017making}, data points that are similar to anchor points will be detected if there are no anchor points available. Then, the cluster-dependent extended transition matrices will be poorly estimated. To handle this problem, we follow \cite{xia2019revision} to revise the cluster-dependent extended transition matrices, which helps lead to a better classifier. We term the systemic proposed method for training a robust classifier against mixed label noise as \textit{Extended T}. In more detail, Extended T-$i$ means that the number of the cluster-dependent transition matrices is $i$. 
\section{Experiments}\label{sec5}
In this section, we first introduce the methods for comparison in the experiments (Section \ref{sec:5.1}). We then introduce the details of the experiments on synthetic datasets (Section \ref{sec:5.2}). The experiments on real-world datasets are finally presented (Section \ref{sec:5.3}). 
\subsection{Comparison methods}\label{sec:5.1}
We compare the proposed method with the following methods: (1) CE, which trains the deep models with the standard cross entropy loss on noisy datasets. (2) GCE \cite{zhang2018generalized}, which handles label noise by exploiting the negative Box-Cox transformation. (3) APL \cite{ma2020normalized}, which combines two robust loss functions which boost each other. (4) DMI \cite{xu2019l_dmi}, which handles label noise from the perspective of the information theory. (5) NLNL \cite{kim2019nlnl}, which proposes a novel learning method called Negative Learning (NL) to reduce the side effect of label noise. (6) Co-teaching \cite{han2018co}, which trains two networks simultaneously and exchanges the selected examples for network updating. (7) Co-teaching+ \cite{yu2019does}, which trains two networks simultaneously and finds confident examples among the prediction disagreement data. (8) JoCor \cite{wei2020combating}, which reduces the diversity of networks to improve the robustness. (9) S2E \cite{yao2020searching}, which utilizes AutoML to handle label noise. (10) Forward \cite{patrini2017making}, which estimates the class-dependent transition matrix to correct the training loss. (11) T-Revision \cite{xia2019revision}, which introduces a slack variable to revise the estimated transition matrix and leads to a better classifier. Note that  we
do not compare with some state-of-the-art methods like SELF \cite{nguyen2020self}, DivideMix \cite{li2020dividemix}, and EvidentialMix \cite{sachdeva2020evidentialmix}. It is because that their proposed methods are aggregations of multiple advanced approachs while this work only focuses on one, therefore the comparison is not fair. To measure the performance, we use the classification accuracy on the test set. Intuitively, higher accuracy means that the method is more robust to the mixed label noise. Note that for a fair comparison, we implement all methods with default parameters by PyTorch, and conduct all the experiments on NVIDIA Tesla V100 GPUs.

\subsection{Experiments on synthetic datasets}\label{sec:5.2}
\subsubsection{Datasets and Implementation details} 
\textbf{Datasets.} We evaluate the robustness of the proposed method to label noise on synthetic CIFAR-10 \cite{krizhevsky2009learning} with comprehensive experiments. Original CIFAR-10 consists of 50,000 training images and 10,000 test images with 10 classes. The size of images is 32$\times$32$\times$3. Note that the original CIFAR-10 contains clean training data. We thus corrupt the training data manually to generate label noise. Specifically, for the open-set label noise, we follow \cite{wang2018iterative} and borrow the images from SVHN \cite{netzer2011svhn}, CIFAR-100 \cite{krizhevsky2009learning}, and ImageNet32 (32$\times$32$\times$3 ImageNet images) \cite{chrabaszcz2017downsampled} to act as outside images. Note that only images whose labels exclude 10 classes in CIFAR-10 are considered. We then use the outside images to replace some training images in CIFAR-10. For closed-set label noise, we consider the symmetric noise, which is more challenging than asymmetric noise \cite{patrini2017making}. Note that we consider controlled 
class-dependent label noise. It is because we can show the estimation results more clearly. The overall label noise rate is denoted by $\tau\in\{0.2,0.4,0.6,0.8\}$. The proportion of open-set noise in the label noise is denoted by $\rho\in\{0.25,0.5,0.75\}$, and the proportion of closed-set noise is $1-\rho$. We leave out 10\% of the training data as a validation set, which is used for model selection. The experimental results are reported over five trials. We conduct signiﬁcance tests to show whether the experimental results are statistically significant. The details for signiﬁcance tests can be found in Appendix A.\\
\textbf{Implementation.} We employ a PreAct ResNet-18 network \cite{he2016identity}. For learning the transition matrix, we follow the optimization method in \cite{patrini2017making,xia2019revision}. We next use SGD optimizer with momentum 0.9, batch size 128, and weight decay 5$\times10^{-4}$ to initialize the network. The initial learning rate is set to $10^{-2}$, and divided by 10 after the 40th epochs and 80th epochs. 100 epochs are set totally. Following \cite{xia2019revision}, we then exploit Adam optimizer with a learning rate 5$\times10^{-7}$ to revise the transition matrix. Typical data augmentations including random crop and horizontal ﬂip are applied. 
\begin{table*}[!t]
	\centering
	\tiny
	\begin{tabular}{lp{10mm}p{0.001mm}|ccc|ccc|ccc|ccc} 
		\toprule	 	
		     &\multirow{2}{*}{Method}&$\tau$& \multicolumn{3}{c|}{0.2}&\multicolumn{3}{c|}{0.4}&\multicolumn{3}{c|}{0.6}&\multicolumn{3}{c}{0.8}\\
			 \cmidrule{4-15}
		     & &$\rho$ & 0.25 & 0.5 & 0.75 & 0.25 & 0.5 & 0.75 & 0.25 & 0.5 & 0.75 & 0.25 & 0.5 & 0.75 \\
			\midrule
			\parbox[t]{0.1mm}{\multirow{12}{*}{\rotatebox{90}{CIFAR-10+SVHN}}}& CE &  & 90.01$\pm$0.12 & 89.82$\pm$0.35&90.18$\pm$0.16&87.82$\pm$0.35&88.20$\pm$0.16&88.27$\pm$0.17&83.18$\pm$0.65&84.91$\pm$0.11&86.44$\pm$0.67&75.07$\pm$0.85&78.44$\pm$0.92&81.72$\pm$0.34\\
			& GCE & &90.71$\pm$0.20&90.56$\pm$0.22&90.88$\pm$0.20&90.04$\pm$0.21&90.06$\pm$0.25&89.54$\pm$0.14& 85.82$\pm$0.22&86.21$\pm$0.26&86.26$\pm$0.16&82.57$\pm$0.76&83.80$\pm$0.28&84.12$\pm$0.29 \\
			& APL & &90.23$\pm$0.17&90.53$\pm$0.14&90.01$\pm$0.38&90.21$\pm$0.25&88.25$\pm$0.29&88.21$\pm$0.43&84.27$\pm$0.39 &84.51$\pm$0.37&84.52$\pm$0.29&82.95$\pm$0.18&82.02$\pm$0.50&82.11$\pm$0.36\\
			& DMI & &90.71$\pm$0.10&90.04$\pm$0.33&90.58$\pm$0.17&89.87$\pm$0.72&89.94$\pm$0.27&89.02$\pm$0.47&85.96$\pm$0.85&85.21$\pm$0.64&85.76$\pm$0.95&81.95$\pm$1.07&82.01$\pm$0.78&80.27$\pm$0.60\\
			& NLNL &&89.85$\pm$0.19&89.72$\pm$0.32&89.98$\pm$0.27&87.39$\pm$0.28&87.80$\pm$0.50&88.78$\pm$0.29&82.90$\pm$0.55 &84.58$\pm$0.41&84.37$\pm$0.72&76.39$\pm$0.59&79.65$\pm$0.54&78.39$\pm$1.08\\
			& Co-teaching& &90.50$\pm$0.07&90.62$\pm$0.15&90.87$\pm$0.13&86.64$\pm$0.25&87.48$\pm$0.40&87.31$\pm$0.28&75.10$\pm$0.48&76.25$\pm$1.08&78.78$\pm$2.48&46.36$\pm$2.22&49.20$\pm$2.89&53.44$\pm$2.34\\
			& Co-teaching+&&88.46$\pm$0.54&89.51$\pm$0.38&88.32$\pm$0.40&86.39$\pm$0.51&86.71$\pm$0.21&84.92$\pm$0.56&63.18$\pm$4.87&65.29$\pm$9.84&56.41$\pm$8.83&10.07$\pm$1.07&17.06$\pm$8.07&15.38$\pm$2.93\\
			& JoCor&&88.25$\pm$0.06&89.15$\pm$0.21&89.15$\pm$0.45&84.16$\pm$1.07&82.12$\pm$1.03&82.02$\pm$0.74&67.29$\pm$1.23&69.02$\pm$1.72&71.70$\pm$1.73&43.93$\pm$0.32&42.82$\pm$1.31&40.12$\pm$3.44    \\
			& S2E&&89.42$\pm$1.35&89.68$\pm$1.13&89.87$\pm$1.80&88.24$\pm$2.48&88.99$\pm$1.94&88.78$\pm$1.57&81.16$\pm$2.49&85.44$\pm$1.62&85.48$\pm$2.32&57.45$\pm$4.17&74.16$\pm$4.52&78.39$\pm$3.46  \\
			& Forward & &89.37$\pm$0.14&89.27$\pm$0.86&89.64$\pm$0.80&87.54$\pm$0.24&87.76$\pm$0.54& 87.01$\pm$0.39&80.19$\pm$2.72&83.21$\pm$0.89&83.92$\pm$1.98&78.05$\pm$2.02&80.32$\pm$1.84&78.66$\pm$1.72\\
			& T-Revision&&90.23$\pm$0.14&89.97$\pm$0.23&90.02$\pm$0.14&88.68$\pm$0.24 &88.79$\pm$0.29&89.02$\pm$0.47&85.07$\pm$1.03&85.37$\pm$1.09&85.42$\pm$0.83&81.04$\pm$2.04&81.36$\pm$0.97&82.98$\pm$1.17\\	
			\midrule			
			& Extended T& &90.86$\pm$0.13&\textbf{90.89$\pm$0.28}&90.78$\pm$0.16&\textbf{90.94$\pm$0.22}&\textbf{90.72$\pm$0.37}&90.68$\pm$0.38&87.34$\pm$0.38& 86.92$\pm$0.93&87.18$\pm$0.75&83.68$\pm$0.47&83.94$\pm$1.02&84.83$\pm$1.42   \\
			& Extended T-2& &\textbf{90.92$\pm$0.08}&90.58$\pm$0.54&\textbf{91.03$\pm$0.22}&90.73$\pm$0.28&90.54$\pm$0.30&90.42$\pm$0.51&87.32$\pm$0.77&87.03$\pm$1.07&87.08$\pm$0.99&84.02$\pm$0.76&84.02$\pm$1.08&84.77$\pm$1.53   \\
			& Extended T-3& &90.89$\pm$0.17&90.77$\pm$0.18&91.02$\pm$0.17&90.67$\pm$0.42&90.67$\pm$0.41&\textbf{90.72$\pm$0.69}&\textbf{87.48$\pm$0.63}&\textbf{87.19$\pm$0.92}&\textbf{87.29$\pm$0.73}&\textbf{84.42$\pm$0.93}&\textbf{84.31$\pm$0.94}&\textbf{84.88$\pm$1.07}   \\

			\midrule
			\midrule
			\parbox[t]{0.1mm}{\multirow{12}{*}{\rotatebox{90}{CIFAR-10+CIFAR-100}}}& CE &  &87.90$\pm$0.38&87.87$\pm$0.33&87.81$\pm$0.21&86.77$\pm$0.35&86.32$\pm$0.29&86.35$\pm$0.49&82.56$\pm$0.83 &83.81$\pm$0.32&83.87$\pm$0.32&75.55$\pm$0.87&77.85$\pm$0.58&78.40$\pm$0.42\\
			& GCE & &89.65$\pm$0.29&89.33$\pm$0.30&89.68$\pm$0.32&\textbf{89.61$\pm$0.27}&88.26$\pm$0.21&88.87$\pm$0.24& 86.16$\pm$0.72&85.80$\pm$0.18&85.64$\pm$0.38&82.75$\pm$0.44&81.38$\pm$0.78&82.47$\pm$0.51 \\
			& APL & &89.01$\pm$0.12&89.01$\pm$0.25&88.42$\pm$0.53&89.23$\pm$0.37&88.14$\pm$0.33&88.23$\pm$0.19&86.75$\pm$0.26 &86.89$\pm$0.20&86.01$\pm$0.36&83.17$\pm$0.40&80.85$\pm$0.38&83.11$\pm$0.30\\
			& DMI & &89.09$\pm$0.17&88.47$\pm$0.86&88.97$\pm$0.45&87.62$\pm$0.64&87.98$\pm$0.39&87.66$\pm$0.48&83.92$\pm$0.79&84.02$\pm$0.96&84.57$\pm$0.76&80.27$\pm$0.62&80.16$\pm$0.99&79.81$\pm$1.38\\
			& NLNL &&89.27$\pm$0.24&89.88$\pm$0.59&90.06$\pm$0.09&87.06$\pm$0.20&87.55$\pm$0.28&87.77$\pm$0.25&83.72$\pm$0.27&84.00$\pm$0.19&84.25$\pm$0.27&78.02$\pm$0.43&79.10$\pm$0.67&81.04$\pm$1.16\\
			& Co-teaching& &90.06$\pm$0.16&90.01$\pm$0.33&90.01$\pm$0.12&86.47$\pm$0.15&87.27$\pm$0.33&87.39$\pm$0.30&75.34$\pm$1.60&76.52$\pm$0.87&76.08$\pm$2.73&47.56$\pm$2.39&44.91$\pm$2.81&42.60$\pm$2.48\\
			& Co-teaching+&&88.28$\pm$0.47&88.48$\pm$0.50&87.52$\pm$0.16&86.34$\pm$0.67&86.18$\pm$0.19&84.25$\pm$0.23&66.59$\pm$6.22&70.97$\pm$9.27&59.27$\pm$9.06&16.77$\pm$5.29&10.87$\pm$1.63&14.26$\pm$3.01\\
			& JoCor&&88.13$\pm$0.20&88.77$\pm$0.21&89.25$\pm$0.14&79.71$\pm$0.73&81.56$\pm$0.95&82.14$\pm$0.15&65.88$\pm$1.19&67.50$\pm$0.90&68.35$\pm$0.62&39.90$\pm$3.62&37.67$\pm$3.24&40.07$\pm$3.23    \\
			& S2E&&89.32$\pm$1.42&89.21$\pm$1.09&89.36$\pm$1.45&88.18$\pm$2.31&87.51$\pm$2.31&86.07$\pm$3.17&79.21$\pm$4.81&81.52$\pm$3.62&80.15$\pm$3.61&54.51$\pm$4.65&66.79$\pm$4.76&66.73$\pm$4.76  \\
			& Forward & &88.25$\pm$0.14&88.03$\pm$0.92&87.10$\pm$0.47&85.21$\pm$0.93&85.22$\pm$1.93&86.23$\pm$0.80&81.30$\pm$0.54&83.98$\pm$0.75&81.77$\pm$1.06&80.02$\pm$2.42&79.68$\pm$2.87&78.05$\pm$3.08\\
			& T-Revision&&88.81$\pm$0.09&89.05$\pm$0.19&89.57$\pm$0.32&86.63$\pm$0.24&87.33$\pm$0.28&87.26$\pm$0.48&85.78$\pm$0.93 &85.63$\pm$1.25&85.36$\pm$1.28&82.49$\pm$0.97&82.20$\pm$1.05&82.27$\pm$1.73\\	
			\midrule			
			& Extended T& &90.27$\pm$0.20&90.34$\pm$0.37&90.03$\pm$0.80&89.52$\pm$0.71&89.32$\pm$0.32&89.42$\pm$0.36&87.33$\pm$0.67&87.26$\pm$0.43&\textbf{87.39$\pm$0.90}&84.06$\pm$1.35&84.07$\pm$1.42&84.62$\pm$1.40   \\
			& Extended T-2& &90.26$\pm$0.18&90.38$\pm$0.40&\textbf{90.17$\pm$0.58}&89.54$\pm$0.68&\textbf{89.73$\pm$0.65}&89.60$\pm$0.74&87.02$\pm$0.50&87.13$\pm$0.32&86.74$\pm$1.56&84.03$\pm$0.72&\textbf{84.10$\pm$0.87}&\textbf{84.74$\pm$0.92}   \\
			& Extended T-3& &\textbf{90.33$\pm$0.25}&\textbf{90.41$\pm$0.39}&90.14$\pm$0.39&89.47$\pm$0.53&89.37$\pm$0.65&\textbf{89.64$\pm$0.45}&\textbf{87.51$\pm$0.72}&\textbf{87.33$\pm$0.70}&87.30$\pm$1.08&\textbf{84.38$\pm$0.92}&83.92$\pm$1.17&84.59$\pm$1.14   \\
			\midrule
			\midrule	 		
			\parbox[t]{0.1mm}{\multirow{12}{*}{\rotatebox{90}{CIFAR-10+ImageNet32}}}& CE &  &90.18$\pm$0.12&90.59$\pm$0.32&90.54$\pm$0.52&87.69$\pm$0.64&88.97$\pm$0.34&89.02$\pm$0.52&82.39$\pm$0.57&84.64$\pm$1.16&86.03$\pm$0.88&75.94$\pm$0.52&79.71$\pm$1.22&81.05$\pm$2.57\\
			& GCE & &90.59$\pm$0.15& 90.63$\pm$0.17&\textbf{90.73$\pm$0.12}&90.01$\pm$0.34&89.98$\pm$0.37&89.20$\pm$0.68&86.07$\pm$0.51&86.25$\pm$0.63&86.37$\pm$0.42&82.92$\pm$1.07&82.03$\pm$1.89&82.36$\pm$0.82 \\
			& APL & &89.02$\pm$0.48&89.37$\pm$0.23&88.65$\pm$1.07&88.92$\pm$0.64&89.07$\pm$0.23&88.65$\pm$0.29&84.64$\pm$0.50 &83.88$\pm$0.79&84.56$\pm$1.01&78.62$\pm$3.46&79.39$\pm$3.87&80.67$\pm$0.86\\
			& DMI & &90.48$\pm$0.09&\textbf{90.77$\pm$0.23}&90.71$\pm$0.17&89.90$\pm$0.19&89.16$\pm$0.32&89.27$\pm$0.37&85.62$\pm$0.83&85.02$\pm$0.77&85.32$\pm$0.80&77.53$\pm$2.08&80.45$\pm$1.39&80.19$\pm$2.68\\
			& NLNL &&89.92$\pm$0.28&89.93$\pm$0.18&89.97$\pm$0.11&87.06$\pm$0.22&87.33$\pm$0.24&87.50$\pm$0.48&83.30$\pm$0.49&84.70$\pm$0.53&85.02$\pm$0.50&76.57$\pm$0.61&79.66$\pm$0.82&80.05$\pm$1.02\\
			& Co-teaching& &90.31$\pm$0.06&90.20$\pm$0.07&90.37$\pm$0.20&86.48$\pm$0.58&87.66$\pm$0.31&88.34$\pm$0.25&75.60$\pm$1.04&77.33$\pm$1.50&78.53$\pm$3.61&45.89$\pm$2.15&48.09$\pm$3.09&46.67$\pm$5.90\\
			& Co-teaching+&&89.38$\pm$0.23&89.36$\pm$0.40&89.01$\pm$0.65&87.90$\pm$0.54&87.82$\pm$0.23& 86.51$\pm$0.38&76.58$\pm$3.33&79.76$\pm$4.21&75.20$\pm$3.00&10.34$\pm$0.47&10.73$\pm$0.84&20.07$\pm$9.62\\
			& JoCor&&88.18$\pm$0.18&88.90$\pm$0.16&89.41$\pm$0.12&80.89$\pm$0.51&82.62$\pm$0.71&83.29$\pm$0.46&67.16$\pm$1.25&68.69$\pm$1.02&71.00$\pm$1.21&43.63$\pm$2.87&40.70$\pm$4.62&38.88$\pm$4.26   \\
			& S2E&&89.43$\pm$0.75&89.63$\pm$1.21&89.53$\pm$0.99&88.72$\pm$2.38&88.72$\pm$1.03&87.70$\pm$2.34&82.24$\pm$5.18&82.71$\pm$2.82&83.10$\pm$1.64&60.08$\pm$7.93&75.43$\pm$4.44&78.84$\pm$3.28\\
			& Forward & &89.34$\pm$0.96&89.33$\pm$0.53&89.14$\pm$0.73&88.92$\pm$0.73&88.53$\pm$0.62&88.32$\pm$0.98&84.58$\pm$0.71&83.49$\pm$0.74&83.77$\pm$0.97&76.02$\pm$3.09&76.82$\pm$3.46&78.89$\pm$4.52\\
			& T-Revision&&89.72$\pm$0.28&90.01$\pm$0.24&89.95$\pm$0.37&89.45$\pm$0.65&89.09$\pm$0.78&89.02$\pm$0.96&85.72$\pm$0.93&85.25$\pm$1.01&85.03$\pm$1.78&81.17$\pm$2.77&81.34$\pm$2.53&81.37$\pm$2.86\\	
			\midrule			
			& Extended T& &90.62$\pm$0.13&90.47$\pm$0.28&90.63$\pm$0.19&90.17$\pm$0.32&90.10$\pm$0.29&90.18$\pm$0.50&87.07$\pm$0.37&87.53$\pm$0.92&87.47$\pm$0.93&84.03$\pm$0.94&84.51$\pm$1.07&84.57$\pm$1.96   \\
			& Extended T-2& &\textbf{90.73$\pm$0.30}&90.60$\pm$0.48&90.68$\pm$0.29& \textbf{90.26$\pm$0.75}&90.29$\pm$0.47&90.23$\pm$1.02&\textbf{87.62$\pm$0.77}&87.58$\pm$0.76&87.42$\pm$1.17&\textbf{84.32$\pm$0.93}&84.42$\pm$1.25&84.60$\pm$1.20   \\
			& Extended T-3& &90.69$\pm$0.26&90.58$\pm$0.18&\textbf{90.73$\pm$0.21}&\textbf{90.26$\pm$0.68}&\textbf{90.34$\pm$0.52}&\textbf{90.31$\pm$1.06}&87.49$\pm$0.82&\textbf{87.62$\pm$1.04}&\textbf{87.57$\pm$1.34}&84.19$\pm$1.18&\textbf{84.52$\pm$0.96}&\textbf{84.64$\pm$1.48}  \\
		\bottomrule
\end{tabular}
	\caption
		{
		Mean and standard deviations of test accuracies (\%) on synthetic noisy datasets with different noise settings. The best experimental results are in bolded. 
		}
	\label{tab:cifar}
\end{table*}
\subsubsection{Classification accuracy}
We report comprehensive experimental results on the synthetic datasets in Table \ref{tab:cifar}. For CIFAR-10+SVHN, we can clearly see that the proposed method consistently outperform the the prior state-of-the-art approaches for learning with mixed label noise. Specially, in the cases of high label noise rates, e.g., $\tau=0.6$ and $\tau=0.8$, our method Extended T-3 achieves the best classification performance. This means that the proposed cluster-dependent transiton matrix can reduce the estimation error brought by randomness and uncertainty in the estimation process, which helps lead to better classifiers. For CIFAR-10+CIFAR-100, our method achieves the superior performance again, and get ahead of the other methods in almost all cases. Note that the baseline GCE achieves the best performance when $\tau=0.4$ and $\rho=0.25$. The reasons may be that the open-set label noise rate and overall label noise rate are both low, and GCE can cope with the symmetric closed-set label noise well in this case. However, when in the face of more challenging cases, our method is better for handling the mixed label noise. For instance, when $\tau=0.8$ and $\rho=0.75$, the open-set label noise rate and overall label noise rate are both high, the proposed Extended T achieves a more than 2\% lead over GCE. Compared with the model-based baselines, e.g., Forward and T-Revision, the proposed method always works better, which means that we can effectively extend to model the mixed label noise and thus obtain better robustness. For CIFAR-10+ImageNet32, in the easier cases, e.g., $\tau=2$, we can see that DMI and GCE perform better than our method in some cases. In more challenging cases, the proposed method again surpasses all baselines. To sum up, the synthetic experiments reveal that our method is powerful in handling mixed label noise, particularly in the case of high noise rates.
\subsubsection{Estimation results}

\begin{figure}[t]
\centering
\subfigure[]{\label{fig2:a} 
\begin{minipage}[t]{0.5\linewidth}
\centering
\includegraphics[width=2.20in]{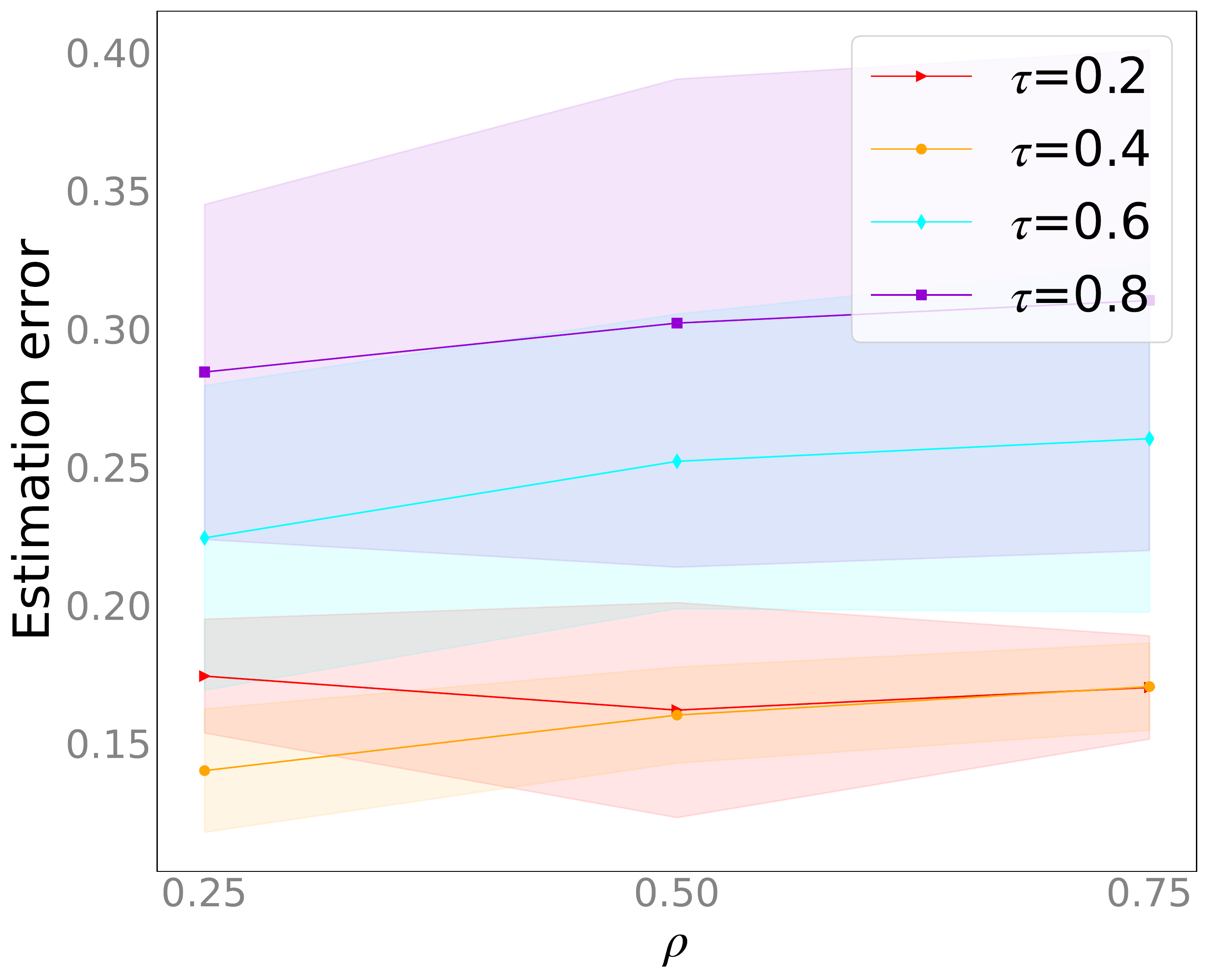}
\label{error}
\end{minipage}
}%
\subfigure[]{\label{fig2:b} 
\begin{minipage}[t]{0.5\linewidth}
\centering
\includegraphics[width=2.20in]{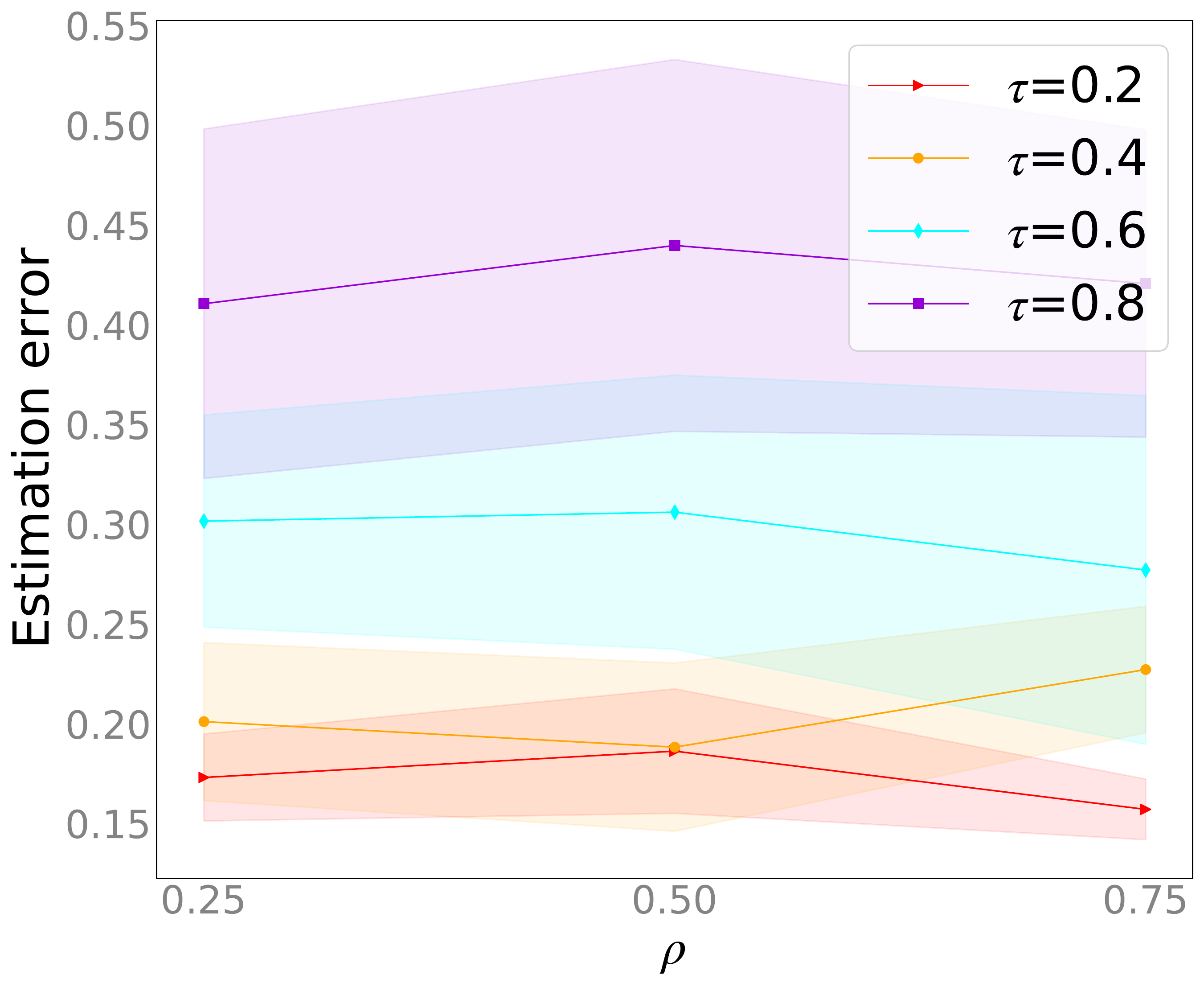}
\end{minipage}
}
\centering
\caption{Illustration of the transition matrix estimation error. Figure (a) illustrates the estimation error for modeling open-set label noise. Figure (b) illustrates the estimation error for modeling mixed label noise. The error bar for standard deviation in each figure has been shaded.}   
\label{fig2}
\end{figure}

 We report the estimation error of the transition matrix $T^{\circ}$ for open-set label noise, and of the transition matrix $T^{\star}$ for mixed label noise. Note that only our method extends the traditional transition matrix to model the open-set label noise and further model the mixed label noise. We thus do not report the estimation results of other model-based methods. The experiments are conducted on CIFAR10+SVHN by exploiting the proposed Extended T. The experimental settings are same as before. The estimation errors are calculated with $\ell_1$ norm. The results are presented in Fig.~\ref{fig2}. 
\subsection{Experiments on real-world face datasets}\label{sec:5.3}
\subsubsection{Datasets}
\textbf{Training data.} The training datasets include VggFace-2 \cite{cao2018vggface2} and MS1MV0 \cite{guo2016ms}. VggFace-2 is a noisy dataset collected from Google image search. MS1MV0 is a raw dataset with a large amount of noisy labels \cite{wang2018devil,deng2020subcenter}.  \\
\textbf{Test data.} We use four popular benchmarks as test datasets, including CFP \cite{sengupta2016frontal}, AgeDB \cite{moschoglou2017agedb}, CALFW \cite{zheng2017cross}, and CPLFW \cite{zheng2018cross}. CFP consist of face images of celebrities in frontal and profile views. AgeDB contains images annotated with accurate to the year, noise-free labels.  CALFW considers a more general cross-age situation and provides a face image set with large intra-class variations. CPLFW is similar to CALFW, but considers a cross-pose case. We summarize the important statistics of the used datasets in Appendix B. 

\subsubsection{Experimental settings}
\textbf{Data processing.} We follow ArcFace \cite{deng2019arcface} to generate the normalised face crops by exploiting five facial points (two eyes, nose tip, and two mouth corners) predicted by RetinaFace \cite{deng2020retinaface}. The size of the face crops is 112$\times$112. \\
\textbf{Network structure and activation function.} To be fair, in this paper, we employ the same architecture and the activation function for testing different baselines. We use MobileFaceNet \cite{chen2018mobilefacenets} and Arc-Softmax \cite{deng2019arcface}, which are popular in the face recognition task.  The dimension of the face embedding feature is 512. For the angular margin $m$ and feature scale $s$, we set 0.5 and 32, respectively. \\
\textbf{Training and testing.} At the train stage, we train the deep models with SGD with momentum 0.9, with total batch size 512 on 4 GPUs parallelly and weight decay $5\times10^{-4}$. For learning an initial classifer, the learning rate is initially 0.1 and divided by 10 at the 5th, 10th, and 15th epochs. We set 20 epochs in total. For learning the classiﬁer and slack variable, Adam is used and the learning rate is changed to $5\times10^{-7}$. At the test stage, we use MobileFaceNet to extract the 512-$D$ feature embeddings of test face images. We follow the unrestricted with labelled outside data protocol \cite{huang2008labeled} to report the verification performance on test face datasets. 
\subsubsection{Experimental results}
\begin{table}[h]
	\centering
	\begin{tabular}{c|cccc|c} 
		\toprule	 	
		    Method&CFP&AgeDB&CALFW&CPLFW&Ave.\\
			\midrule
			CE&95.30&92.69&89.94&85.97&90.98\\
			GCE&94.26&91.06&89.98&85.28&90.15\\
			APL&94.16&90.29&88.32&85.09&89.47\\
			DMI&94.39&92.83&90.06&85.86&90.79\\
			NLNL&86.74&87.63&84.79&80.06&84.81\\
			Co-teaching&95.47&92.53&89.58&85.32&90.73\\
			Co-teaching+&95.26&92.01&85.12&85.19&89.40\\
		    JoCor&92.32&90.97&84.58&82.11&87.50\\
		    S2E&92.09&91.64&89.70&84.63&89.52\\
			Forward&95.07&92.40&89.10&85.79&90.59\\
			T-Revision&95.38&92.79&89.86&85.94&90.99\\
			\midrule	 		
		    Extended T&95.57&93.06&90.33&86.24&91.30\\
		    Extended T-2&95.59&\textbf{93.15}&90.42&86.36&91.38\\
		    Extended T-3&\textbf{95.73}&93.14&\textbf{90.67}&\textbf{86.52}&\textbf{91.51}\\
		\bottomrule
\end{tabular}
	\caption
		{
		Test accuracies (\%) of different methods training on VggFace-2. The best results are in bolded. 
		}
	\label{tab:vggface2}
\end{table}			

\begin{table}[h]
	\centering
	\begin{tabular}{c|cccc|c} 
		\toprule	 	
		    Method&CFP&AgeDB&CALFW&CPLFW&Ave.\\
			\midrule
			CE&88.79&92.35&91.36&82.92&88.86\\
			GCE&89.02&91.87&91.32&82.77&88.75\\
			APL&87.06&91.55&91.29&82.02&87.98\\
			DMI&89.02&92.18&90.94&83.01&88.79\\
			NLNL&83.06&85.47&84.21&74.02&81.69\\
			Co-teaching&91.25&93.05&91.24&84.02&89.89\\
			Co-teaching+&91.36&91.87&90.93&84.53&89.67\\
		    JoCor&86.16&89.30&88.09&79.84&85.85\\
		    S2E&91.52&91.61&91.03&81.47&88.91\\
			Forward&90.02&92.32&91.28&82.95&89.14\\
			T-Revision&90.29&92.49&91.60&83.82&89.55\\
			\midrule	 		
		    Extended T&90.33&92.59&91.65&83.91&89.62\\
		    Extended T-2&91.65&93.05&92.30&84.67&90.42\\
		    Extended T-3&\textbf{92.08}&\textbf{93.71}&\textbf{92.61}&\textbf{85.31}&\textbf{90.93}\\
		\bottomrule
\end{tabular}
	\caption
		{
		Test accuracies (\%) of different methods training on MS1MV0. The best results are in bolded. 
		}
	\label{tab:ms1mv0ab}
\end{table}	

We use two training datasets, i.e., VggFace-2 and MS1MV0, to separately train the deep networks. In Table \ref{tab:vggface2} and \ref{tab:ms1mv0ab}, we show the results of the proposed method and baselines on CFP \cite{sengupta2016frontal}, AgeDB \cite{moschoglou2017agedb}, CALFW \cite{zheng2017cross}, and CPLFW \cite{zheng2018cross}, respectively. We can observe that the most of the results obtained by training on VggFace-2 are higher than the results on MS1MV0. It is because that MS1MV0 contains more noisy labels, and therefore is more challenging \cite{deng2020subcenter,wang2019co}. To our method, we can effectively model the mixed label noise, and thus can achieve higher performance than the baselines. Specifically, when training on VggFace-2, Extended T consistently outperforms the baselines. Note that NLNL achieve poor performance in most cases, it is because NLNL exploits negative learning and choose the complementary labels to avoid wrong information brought by noisy labels. Real-world datasets are more complex, though NLNL aims to avoid wrong information, it is hard to select the possible clean labels included in complementary labels. The improvement of classification performance brought by changing the number of cluster is not too obvious. It is because that the label noise rate of VggFace-2 is low \cite{wang2019co}. When training on MS1MV0, we can see that the proposed method surpass the baselines again. Specifically, compared with Forward and T-Revision, the proposed method leads them clearly. Compared with the methods which empirically work well, the proposed method still outperforms them. It is worth noting that the cluster-dependent transition matrices brings a significant performance improvement. The improvement shows that the cluster-dependent transition matrices can model the complex label noise more accurately in this realistic scenario.
\section{Conclusion}\label{sec6}
In this paper, we investigate into learning with mixed closed-set and open-set noisy labels, which is more practical but lacks systematic study in current works. We extend the traditional transition matrix to be able to model mixed label noise, and exploit the cluster-dependent extended transition matrices to better approximate the instance-dependent label noise. Empirical evaluations on synthetic and real-world datasets show the effectiveness of the proposed method for modeling label noise and leading to better classifiers. We believe that our work will urge the research community to explore the robustness of algorithms in this realistic noisy label scenario. 


\newpage
\bibliography{bib}
\newpage
\appendix
\section{The details of significance tests}
We exploit significance tests to show whether all experimental results are statistically significant. The \textit{p}-values are obtained with the two independent samples t-test \cite{sedgwick2010independent}. Note that small \textit{p}-values reflect the performance of the proposed method is significantly better than the performance of the baselines. The significance test results of the baselines compared with our method are presented in Table \ref{tab:tests}. We can see that almost all results are statistically significant.
\begin{table*}[!h]
	\centering
	\footnotesize
	\begin{tabular}{lp{17mm}p{0.001mm}|ccc|ccc|ccc|ccc} 
		\toprule	 	
		     &\multirow{2}{*}{Method}&$\tau$& \multicolumn{3}{c|}{0.2}&\multicolumn{3}{c|}{0.4}&\multicolumn{3}{c|}{0.6}&\multicolumn{3}{c}{0.8}\\
			 \cmidrule{4-15}
		     & &$\rho$ & 0.25 & 0.5 & 0.75 & 0.25 & 0.5 & 0.75 & 0.25 & 0.5 & 0.75 & 0.25 & 0.5 & 0.75 \\
			\midrule
			\parbox[t]{0.1mm}{\multirow{12}{*}{\rotatebox{90}{CIFAR-10+SVHN}}}& CE &  &0.0000&0.0016&0.0007&0.0000&0.0000&0.0000&0.0000&0.0120&0.1798&0.0000&0.0000&0.0103\\
			& GCE & &0.2351&0.1030&0.4584&0.0004&0.0211&0.0023& 0.0003&0.2060&0.0691&0.0436&0.8027&0.3787 \\
			& APL & &0.0005&0.0620&0.0118&0.0024&0.0000&0.0000&0.0000 &0.0043&0.0010&0.0326&0.0150&0.0166\\
			& DMI & &0.0786&0.0046&0.1257&0.0384&0.0106&0.0007&0.0277&0.0188&0.0486&0.0281&0.0183&0.0015\\
			& NLNL &&0.0001&0.0006&0.0018&0.0000&0.0000&0.0001&0.0000 &0.0046&0.0006&0.0000&0.0003&0.0001\\
			& Co-teaching& &0.0007&0.1391&0.4091&0.0000&0.0004&0.0000&0.0000&0.0000&0.0016&0.0000&0.0000&0.0000\\
			& Co-teaching+&&0.0000&0.0005&0.0001&0.0000&0.0000&0.0000&0.0006&0.0114&0.0021&0.0000&0.0001&0.0000\\
			& JoCor&&0.0000&0.0000&0.0010&0.0001&0.0000&0.0000&0.0000&0.0000&0.0000&0.0000&0.0000&0.0000    \\
			& S2E&&0.0997&0.0986&0.3700&0.0949&0.1498&0.0715&0.0110&0.1612&0.2240&0.0002&0.0109&0.0167  \\
			& Forward & &0.0000&0.0167&0.0450&0.0000&0.0000& 0.0000&0.0058&0.0004&0.0266&0.0041&0.0129&0.0006\\
			& T-Revision&&0.0001&0.0011&0.0001&0.0000 &0.0000&0.0007&0.0008&0.0633&0.0138&0.0593&0.0064&0.0805\\	
			\midrule
			\midrule
			\parbox[t]{0.1mm}{\multirow{12}{*}{\rotatebox{90}{CIFAR-10+CIFAR-100}}}& CE &  &0.0000&0.0000&0.0000&0.0005&0.0000&0.0000&0.0000&0.0000&0.0007&0.0000&0.0000&0.0005\\
			& GCE & &0.0095&0.0031&0.0277&0.8219&0.0009&0.0387&0.0448&0.0017&0.0140&0.1262&0.0018&0.0340 \\
			& APL & &0.0000&0.0009&0.0005&0.4960&0.0009&0.0011&0.1654&0.1727&0.0340&0.2581&0.0007&0.0968\\
			& DMI & &0.0000&0.0080&0.0017&0.0043&0.0008&0.0005&0.0002&0.0011&0.0015&0.0027&0.0004&0.0012\\
			& NLNL &&0.0002&0.2297&0.2080&0.0015&0.0001&0.0000&0.0004&0.0000&0.0000&0.0007&0.0036&0.0046\\
			& Co-teaching& &0.1415&0.2203&0.1525&0.0007&0.0000&0.0000&0.0000&0.0000&0.0006&0.0000&0.0000&0.0000\\
			& Co-teaching+&&0.0004&0.0005&0.0000&0.0002&0.0000&0.0000&0.0025&0.0245&0.0000&0.0000&0.0000&0.0000\\
			& JoCor&&0.0000&0.0025&0.0025&0.0000&0.0000&0.0000&0.0000&0.0000&0.0000&0.0000&0.0000&0.0000    \\
			& S2E&&0.2533&0.1079&0.2529&0.3204&0.1929&0.1019&0.0271&0.0322&0.0141&0.0000&0.0016&0.0010  \\
			& Forward & &0.0000&0.0049&0.0000&0.0000&0.0123&0.0005&0.0000&0.0002&0.0000&0.0255&0.0350&0.0000\\
			& T-Revision&&0.0000&0.0761&0.0140&0.0006&0.0000&0.0001&0.0294&0.0575&0.0438&0.0993&0.0262&0.0692\\	
			\midrule
			\midrule	 		
			\parbox[t]{0.1mm}{\multirow{12}{*}{\rotatebox{90}{CIFAR-10+ImageNet32}}}& CE &  &0.0011&0.5882&0.7581&0.0005&0.0011&0.0123&0.0000&0.0036&0.1129&0.0000&0.0004&0.0634\\
			& GCE & &0.7703&0.3631&0.4041&0.5125&0.6242&0.0515&0.0148&0.0549&0.0780&0.1583&0.0603&0.0883\\
			& APL & &0.0019&0.0003&0.0197&0.0133&0.0006&0.0015&0.0001&0.0003&0.0029&0.0329&0.0553&0.0126\\
			& DMI & &0.1192&0.1378&0.5480&0.1932&0.0025&0.0209&0.0211&0.0033&0.0083&0.0016&0.0020&0.0322\\
			& NLNL &&0.0046&0.0147&0.0007&0.0000&0.0000&0.0001&0.0000&0.0015&0.0033&0.0000&0.0001&0.0064\\
			& Co-teaching& &0.0057&0.1267&0.0963&0.0000&0.0000&0.0006&0.0000&0.0000&0.0063&0.0000&0.0000&0.0001\\
			& Co-teaching+&&0.0002&0.0025&0.0059&0.0002&0.0000&0.0000&0.0031&0.0194&0.0007&0.0000&0.0000&0.0000\\
			& JoCor&&0.0000&0.0000&0.0059&0.0000&0.0000&0.0000&0.0000&0.0000&0.0000&0.0000&0.0000&0.0000   \\
			& S2E&&0.0326&0.2411&0.0898&0.2916&0.0532&0.1011&0.1357&0.0238&0.0031&0.0036&0.0000&0.0217\\
			& Forward & &0.0553&0.0087&0.0131&0.0227&0.0043&0.0150&0.0008&0.0002&0.0006&0.0049&0.0133&0.0649\\
			& T-Revision&&0.0041&0.0379&0.0172&0.0341&0.0587&0.0756&0.0409&0.0104&0.0510&0.1090&0.0653&0.1069\\	
		\bottomrule
\end{tabular}
	\caption
		{
		The results of significant tests ($p$-values) on synthetic noisy datasets with different noise settings.
		}
	\label{tab:tests}
\end{table*}

\section{The details of used face image datasets}
The important statistics of the used datasets are summarized in Table \ref{tab:face_datasets}.

\begin{table}[h]
    \centering
    \begin{tabular}{|c|c|c|c|}
    \hline
    &Datasets&\#Identities&\#Images\\
    \hline
    \hline
    \multirow{2}{*}{Train}&VggFace-2 \cite{cao2018vggface2}&9.1K&3.3M \\
    &MS1MV0 \cite{guo2016ms}&100K&10M\\
    \hline
    \multirow{4}{*}{Test}&CFP \cite{sengupta2016frontal}&500&7,000\\
    &AgeDB \cite{moschoglou2017agedb}&568&16,488\\
    &CALFW \cite{zheng2017cross}&5,749&12,174\\
    &CPLFW \cite{zheng2018cross}&5,749&11,652\\
    \hline
    \end{tabular}
    \caption{Face datasets for training and testing.}
    \label{tab:face_datasets}
\end{table}

\end{document}